\def\year{2022}\relax
\documentclass[letterpaper]{article} 
\usepackage{aaai22}  
\usepackage{times}  
\usepackage{helvet}  
\usepackage{courier}  
\usepackage[hyphens]{url}  
\usepackage{graphicx} 
\def\UrlFont{\rm}  
\usepackage{natbib}  
\usepackage{caption} 
\DeclareCaptionStyle{ruled}{labelfont=normalfont,labelsep=colon,strut=off} 
\usepackage{algorithm}
\usepackage{algorithmic}
\usepackage{dsfont}

\usepackage{newfloat}
\usepackage{listings}

\usepackage{times}
\usepackage{epsfig}
\usepackage{graphicx}
\usepackage{amsmath}
\usepackage{amssymb}
\usepackage{dsfont}
\usepackage{booktabs}
\usepackage{xcolor}
\usepackage{xspace}

\usepackage{bm} 
\floatstyle{ruled}
\newfloat{listing}{tb}{lst}{}
\floatname{listing}{Listing}
\title{Barely-Supervised Learning: \\Semi-Supervised Learning with very few labeled images}
\author{
   Thomas Lucas\textsuperscript{\rm 1},
   Philippe Weinzaepfel\textsuperscript{\rm 1},
   Gregory Rogez\textsuperscript{\rm 1}
}
\affiliations{
    \textsuperscript{\rm 1}Naver Labs Europe\footnote{ www.europe.naverlabs.com}\\

    [surname].[lastname]@naverlabs.com
}

\makeatletter
\DeclareRobustCommand\onedot{\futurelet\@let@token\@onedot}
\def\@onedot{\ifx\@let@token.\else.\null\fi\xspace}

\def\eg{\emph{e.g}\onedot} 
\def\ie{\emph{i.e}\onedot} 
 
 \def\vs{\emph{vs}\onedot}
 
\def\etal{\emph{et al}\onedot}
\makeatother

\newcommand\bx{\bm{x}}

\DeclareMathOperator*{\argmax}{arg\,max}

\usepackage{enumitem}
\usepackage{array}
\usepackage{multirow}
\usepackage{stmaryrd}

\newcommand{\camblue}[1]{#1}

\newcommand{\blue}[1]{#1}

\newcommand{\mvspace}[1]{\vspace*{0cm}}

\newcommand{\mypar}[1]{\vspace{0.2cm}\noindent\textbf{#1}}

\newcommand{\tabours}{LESS}

\begin{document}

\maketitle
\begin{abstract}
This paper tackles the problem of semi-supervised learning when the set of labeled samples is limited to a small number of images per class, typically less than 10, problem that we refer to as barely-supervised learning.
We analyze in depth the behavior of a state-of-the-art semi-supervised method, FixMatch, which relies on a weakly-augmented version of an image to obtain supervision signal for a more strongly-augmented version. 
We show that it frequently fails in barely-supervised scenarios, due to a lack of training signal when no pseudo-label can be predicted with high confidence. We propose a method to leverage self-supervised methods that provides training signal in the absence of confident pseudo-labels.
We then propose two methods to refine the pseudo-label selection process which lead to further improvements. The first one relies on a per-sample history of the model predictions, akin to a voting scheme.
The second iteratively updates class-dependent confidence thresholds to better explore classes that are under-represented in the pseudo-labels.
Our experiments show that our approach performs significantly better on STL-10 in the barely-supervised regime, \eg with 4 or 8 labeled images per class. 

\end{abstract}

\section{Introduction}
\label{sec:intro}
While early deep learning methods have reached outstanding performance in 
fully-supervised settings \cite{alexnet,vgg,resnet},
a recent trend is to focus on reducing this need for labeled data.
Self-supervised models take it to the extreme by learning models without any labels; in particular recent works based on the paradigm of contrastive learning~\cite{moco,simclr,byol,swav}, learn features that are invariant to class-preserving augmentations and have shown transfer performances that sometimes surpass that of models pretrained on ImageNet with label supervision.
In practice, however, labels are still required for the transfer to the final task. 
Semi-supervised learning aims at reducing the need for labeled data in the final task, by leveraging both a small set of labeled samples and a larger set of unlabeled samples from the target classes. In this paper, we study the case of semi-supervised learning when the set of labeled samples is reduced to a very small number, typically 4 or 8 per class, which we refer to as \emph{barely-supervised learning}. 

The recently proposed FixMatch approach~\cite{fixmatch} unifies two trends in semi-supervised learning: pseudo-labeling~\cite{pseudolabel} and consistency regularization~\cite{bachman2014learning,rasmus2015semi}.
Pseudo-labeling, also referred to as self-training, consists in accepting confident model predictions as targets for previously unlabeled images, as if they were true labels.
Consistency regularization methods obtain training signal using a modified version of an input, \eg using another augmentation, or a modified version of the model being trained. In FixMatch, a weakly-augmented version of an unlabeled image is used to obtain a pseudo-label as distillation target for a strongly-augmented version of this same image. In practice, the pseudo-label is only set if the prediction is confident enough, as measured by the peakiness of the softmax predictions. If no confident prediction can be made, no loss is applied to the image sample. FixMatch obtains state-of-the-art semi-supervised results, and was the first to demonstrate performance in barely-supervised learning close to fully-supervised methods on  CIFAR-10.
However, it does not perform as well with more realistic images, \eg on the STL-10 dataset when the set of labeled images is small.

\begin{figure*}
\mvspace{-0.65cm}
\centering
\includegraphics[width=\linewidth]{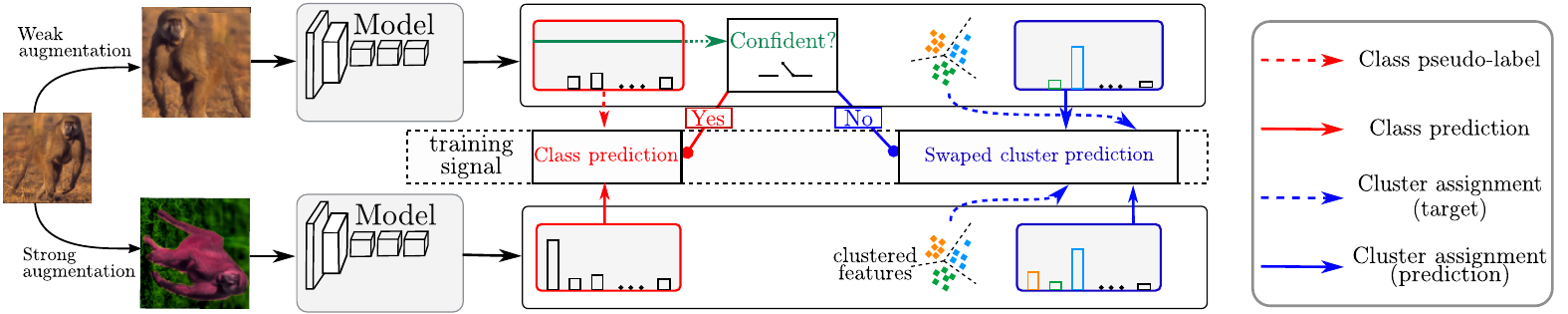}
\caption{Overview of our approach. Given one weak and one strong augmentation of one unlabeled image, we look at the prediction of the weak augmentation. If it is confident, we use it to obtain a pseudo-label which is used for supervision of the strong augmentation. Otherwise, we use feature cluster assignments as target while FixMatch discards the data point.}
\label{fig:overview}
\mvspace{-0.45cm}
\end{figure*}

In this paper, we first analyze the causes that hurt performance in this regime.
In practice, we find the choice of confidence threshold, beyond which a prediction is accepted as pseudo-label, to have a high impact on performance.
A high threshold leads to pseudo-labels that are more likely to be correct, but also to fewer unlabeled images being considered. Thus in practice a smaller subset of the unlabeled data receives training signal, and the model may not be able to make high quality predictions outside of it. If the threshold is set too low, many images will receive pseudo-labels but with the risk of using wrong labels, that may then propagate to other images, a problem known as confirmation bias~\cite{confirmationbias}. In other words, FixMatch faces a \textit{distillation dilemma} between allowing more exploration but with possibly noisy labels, or exploring fewer images with more chances to have correct pseudo-labels.

For barely-supervised learning, a possibility is to leverage a \textit{self-then-semi} paradigm, \ie, to first train a model with self-supervision then with semi-supervised learning, as proposed in SelfMatch~\cite{selfmatch}. We find that this might not be optimal as the self-supervision step ignores the availability of labels for some images. Empirically, we observe that such models tend to output overconfident pseudo-labels in early training, including for incorrect predictions. 

In this paper, \blue{we propose a simple solution to unify self- and semi-supervised strategies, mainly by using a self-supervision signal in cases where no pseudo-label can be assigned with high confidence, see Figure~\ref{fig:overview} for an overview}. 
\blue{Specifically, we perform online deep-clustering ~\cite{swav} and enforce consistency between predicted cluster assignments for two augmented versions of an image when pseudo-labels are not confident.}
This simple algorithmic change leads to clear empirical benefits for barely-supervised learning, owing to the fact that training signal is available even when no pseudo-label is assigned. 
We further propose two strategies to refine pseudo-label selection: (a) by leveraging the history of the model prediction per sample and (b) by imposing constraints on the ratio of pseudo-labeled samples per class. We refer to our method as \textbf{LESS}, for label-efficient semi-supervision.
Our experiments demonstrate substantial benefits from using our approach on STL-10 in barely-supervised settings. For instance, 
test accuracy increases from 35.8\% to 64.2\% when considering 4 labeled images per class, compared to FixMatch. We also improve over other baselines that employ self-supervised learning as pretraining, followed by FixMatch.\\
\textbf{Summary of our main contributions:}
\begin{itemize}[noitemsep,topsep=0pt]

    \item An analysis of the distillation dilemma in FixMatch. We show that it leads to failures with very few labels.
    \item A semi-supervised learning method
    which provides training signal
    in the absence of pseudo-labels and two methods to refine the quality of pseudo-labels.
    \item Experiments showing that our approach allows barely-supervised learning on the more realistic STL-10 dataset.
\end{itemize}
\noindent This paper starts with a review of the related work (Section~\ref{sec:related}), then we analyze the distillation dilemma of FixMatch in Section~\ref{sec:fixmatch}. We propose our method for bare supervision in Section~\ref{sec:method} and our experimental results in Section~\ref{sec:xp}.

\section{Related work}
\label{sec:related}

In this section, we first briefly review related work on semi-supervised learning (Section~\ref{sub:related_semi}) and self-supervised learning (Section~\ref{sub:related_self}). Section~\ref{sub:related} finally discusses recent works that leverage both self- and semi-supervision.

\subsection{Semi-Supervised Learning}
\label{sub:related_semi}

Self-training is a popular method for semi-supervised learning where model predictions are used to provide training signal for unlabeled data, see \cite{selftraining,pseudolabel,selfdistillation}.
In particular, Pseudo-labeling~\cite{pseudolabel} generates artificial labels in the form of hard assignments, typically when a given measure of model confidence, such as the peakiness of the predicted probability distribution, is above a certain threshold~\cite{rosenberg2005semi}. Note that this results in the absence of training signal when no confident prediction can be made. \blue{In \cite{mpl}, a teacher network is trained with reinforcement learning to provide a student network with pseudo-labels that improve its performance}. 
Consistency regularization~\cite{bachman2014learning,rasmus2015semi,sajjadi2016regularization} is based on the assumption that model predictions should not be sensitive to perturbations applied on the input samples. Several predictions are considered for a given data sample, for instance using multiple augmentations or different versions of the trained model.  Artificial targets are then provided by enforcing consistency across these different outputs. This objective can be used as a regularizer, computed on the unlabeled data along with a supervised objective. 

 ReMixMatch~\cite{remixmatch} and Unsupervised Data Augmentation~\cite{uda} (UDA) have shown impressive results by using model predictions on weakly-augmented version of an image to generate artificial target probability distributions. These distributions are then sharpened and used as supervision for a strongly-augmented version of the same image. FixMatch~\cite{fixmatch} provides a simplified version where pseudo-labeling is used instead of distribution sharpening, without the need for additional tricks such as distribution alignment or augmentation anchoring (\ie, using more than one weak and one strong augmented version) from ReMixMatch or training signal annealing from UDA. \blue{Additionally, similar unlabeled images can be encouraged to have consistent pseudo-labels \cite{simple}}, \camblue{or pseudo-labels can be propagated via a similarity graph \cite{dna} or centroids \cite{usadtm}}. 
Our method extends FixMatch by leveraging a self-supervised loss in cases where the pseudo-label is unconfident, allowing to perform barely-supervised learning in realistic settings.

\subsection{Self-Supervised Learning}
\label{sub:related_self}

Early works such as
\cite{doersch2015unsupervised,dosovitskiy2014discriminative,rotnet,jigsaw} on self-supervised learning were based on the idea that a network could learn important image features and semantic representation of the scenes when trained to predict basic transformations applied to the input data, such as a simple rotation in RotNet~\cite{rotnet} or solving a jigsaw puzzle of an image ~\cite{jigsaw}, \ie, recovering the original position of the different pieces.
More recently, impressive results have been obtained using contrastive learning~\cite{instancediscrimination,cmc,simclr,mocov2}, to the point of outperforming supervised pretraining for tasks such as object detection, at least when performing the self-supervision on object-centric datasets~\cite{purushwalkam2020demystifying} such as ImageNet.
The main idea consists in learning feature invariance to class-preserving augmentations. More precisely, each batch contains multiple augmentations of a set of images and the network should output features that are close for variants of a same image and far from those from the other images. In other words, it corresponds to learning instance discrimination, and is closely related to consistency regularization.
Reviewing the literature on this topic is beyond the scope of this paper. Major directions consist in studying the type of augmentation being performed~\cite{asano2019critical,gontijo2020affinity,tian2020makes}, \blue{adapting batch normalisation statistics~\cite{eman}}, the way to provide hard negatives with for instance a queue~\cite{moco,mocov2} or large batch size~\cite{simclr}, or even questioning the need for these negatives~\cite{chen2020exploring,byol}.
\blue{With online deep-clustering~\cite{swav}, the feature invariance principle is slightly relaxed by learning to predict cluster assignments, \ie, encouraging features of different augmentations of an image to be assigned to the same cluster, but not necessarily to be exactly similar}.

\subsection{Combination of self-supervised and semi-supervised learning}
\label{sub:related}

In SelfMatch~\cite{selfmatch}, the authors propose to apply a state-of-the-art semi-supervised method (FixMatch) starting from a model pretrained with self-supervision using SimCLR~\cite{simclr}. Similarly, CoMatch~\cite{comatch} shows that using such a model for initialization performs slightly better than using a randomly initialized network, and \camblue{\cite{boostssl} alternate between self- and semi-supervised training.} In this paper, we depart from the sequential approach of doing self-supervision followed by semi-supervision, with a tighter connection between the two concepts, and empirically demonstrate that it leads to improved performance.
Chen \etal~\citeyearpar{bigself} have proposed another strategy where the self-supervision is first applied, then a classifier is learned on the labeled samples only, which is used to assign a pseudo-label to each unlabeled sample. These pseudo-labels are finally used for training a classifier on all samples. While impressive results are shown on ImageNet with 1\% of the training data, it still represents about 13,000 labeled samples, and may generalize less when considering a lower number of labeled examples.
S4L~\cite{s4l} used a multi-task loss where a self-supervised loss is applied to all samples while a supervised loss is additionally applied to labeled samples only. Similarly to~\cite{bigself}, the classifier is only learned on the labeled samples, a scenario which would fail in the regime of \emph{bare supervision} where very few labeled samples are considered.



\section{The distillation dilemma in FixMatch}
\label{sec:fixmatch}

In this section, we first introduce FixMatch in more details (Section~\ref{sub:fixmatch}) and then formalize the dilemma between exploration \vs pseudo-label accuracy (Section~\ref{sub:dillema}).

\subsection{The FixMatch method}
\label{sub:fixmatch}

\newcommand\mL{\mathcal{L}}
        Let $\mathcal{S} = \{(\bx_i, y_i) \}_{i=1, \hdots M_s}$ be a set of labeled data, sampled from $P_{\bx,y}$.
		In fully-supervised training, the end goal is to learn the optimal parameter $\theta^*$ for a model $p_\theta$, trained to maximize the log-likelihood of predicting the correct ground-truth target, $p_\theta(y | \bx)$,  given the input $\bx$:
	\begin{equation}
	\theta^* = \argmax_\theta \underset{(\bx,y) \sim P_{\bx,y}}{\mathbb{E}} \Big[\log p_\theta(y | \bx)\Big].
	\end{equation}
		In semi-supervised learning, an additional unlabeled set $\mathcal{U} = \{(\bx_j) \}_{j=1, \hdots M_u}$, where $y$ is not observed, can be leveraged. 
        
        \mypar{Self-training} \cite{yarowsky-1995-unsupervised}
	 exploits unlabeled data points using model outputs as targets. Specifically, class predictions with enough probability mass (over a threshold $\tau$) are considered confident and converted to one-hot targets, called \emph{pseudo-labels}. Denote the stop-gradient operator $\bar{f}$, $\hat{y}_x = \argmax(\bm{\bar{p}_\theta}(\bx))$ and $\llbracket \cdot \rrbracket$ the Iverson bracket gives: 
	\begin{equation}
		\underset{\theta}{\text{maximize}} \:\: \underset{\bx \sim P_x}{\mathbb{E}} \big[\: \llbracket \max \bm{\bar{p}_\theta}(\bx) \ge \tau \rrbracket \cdot \log p_\theta(\hat{y}_x|\bx)\: \big].
	\end{equation}
	Ideally, labels should progressively propagate to all $\bx \in \mathcal{U}$.
    
	\mypar{Consistency regularization} is another paradigm which assumes a family of data augmentations $\mathcal{A}$ that leaves the model target unchanged. Denote by $f_\theta(\bx)$ a feature vector, possibly different from $p_\theta$, \eg produced by an intermediate layer of the network. The features produced for two augmentations of the same image are optimized to be similar, as measured by some function $\mathcal{D}$. Let $(v, w) \in \mathcal{A}^2$ and denote $\bm{x_v} = v(\bx)$, the objective can be written:
	\begin{equation}
		 \mathcal{L}_{\text{coreg}}^\theta(\bx_w, \bx_v) = \: 
		 \mathcal{D}\big[f_\theta(\bx_v), f_\theta(\bx_w)\big].
		 \label{eq:cr}
	\end{equation}
		This problem admits constant functions as trivial solutions; numerous methods exist to ensure that relevant information is retained \cite{instancediscrimination,simclr,swav,byol,moco,cmc}.
		
		\mypar{FixMatch.} In the FixMatch algorithm, self-training and consistency-regularization coalesce in a single training loss.
		\emph{Weak} augmentations $w \sim \mathcal{A}_{\text{weak}}$ are applied to unlabeled images, confident predictions are kept as pseudo-labels and compared with model predictions on a \emph{strongly} augmented variant of the image, using $s \sim \mathcal{A}_{\text{strong}}$:
        \mvspace{-0.1cm}
		\begin{equation}
			\hspace*{-0.1cm} \mathcal{L}_{\text{distill}}^{\theta}(\bx_w, \bx_s) =  \llbracket \max \bm{\bar{p}_\theta}(\bm{x_w}) \ge  \tau \rrbracket \cdot \log p_\theta(\hat{y}_{x_w}|\bm{x_s}).
			        \mvspace{-0.15cm}
			\label{eq:distill}
		\end{equation}

\subsection{Formalizing the distillation dilemma}
\label{sub:dillema}

The FixMatch algorithm \cite{fixmatch} has proven successful in learning an image classifier with bare supervision on CIFAR-10.
As we show experimentally, it is not straightforward  to replicate such performance on more challenging datasets such as STL-10.
We now formalize the failure regimes of the FixMatch method.

\begin{figure}
 \centering
  \includegraphics[width=\linewidth]{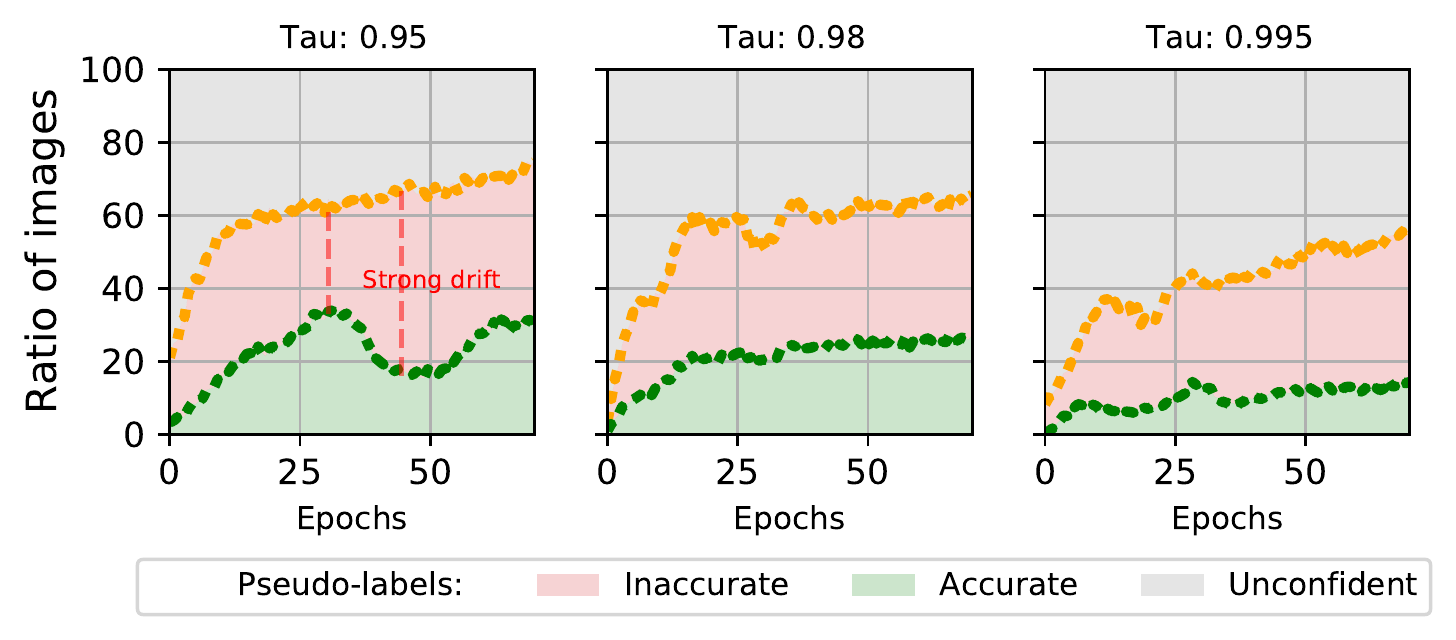} \\[-0.2cm]
  \caption{ Illustration of the distillation dilemma of FixMatch when training on STL-10 with $40$ labels during the first $70$ epochs. For three different values of the confidence threshold $\tau$ (0.95, 0.98 and 0.995 from left to right), we show the ratio of images with a correct and confident pseudo-label (green area in bottom), with an incorrect but confident pseudo-label (red area in middle) and with unconfident pseudo-label (gray area on top) for which no training signal is used. A large value of $\tau$ leads to too few images having a pseudo-label. A lower value allows to leverage more images, but many pseudo-labels are wrong, which is emphasized in later iterations (highlighted between vertical dashed red lines for $\tau=0.95$).}
  \label{fig:dilemma}
  \mvspace{-0.4cm}
\end{figure}

\mypar{Error drift.} Assume model $p_\theta$ is trained with the loss in Eq.~\ref{eq:distill}, and consider the event $E_\theta(\bx, \tau)$ defined as: `the model $p_\theta$ confidently make an erroneous prediction on $\bx$ with confidence threshold $\tau$', then $P(E_\theta(\bx, \tau))$ is equal to :
\begin{equation}
\underset{w \sim \mathcal{A}_{\text{weak}}}{\mathbb{E}} \big[ \llbracket \max \bm{\bar{p}_{\theta}}(\bx_w) \ge \tau \rrbracket \cdot \llbracket \argmax \bm{\bar{p}_{\theta}}(\bx_w) \neq y \rrbracket \big].
\end{equation}
For fixed model parameters $\theta$, $P(E_\theta(\bx, \tau))$ is monotonously decreasing in $\tau$. 
Denote $\theta(t)$ the model parameters at iteration $t$; If the event $E_{\theta(t)}(\bx, \tau)$ occurs at time $t$, by definition optimizing Equation~\ref{eq:distill} leads in expectation to $P(E_{\theta(t+1)}(\bx, \tau)) \ge P(E_{\theta(t)}(\bx, \tau)).$ Thus the model becomes more likely to make the same mistake. Once the erroneous label is accepted, it can propagate to data points similar to $\bx$, as happens with ground-truth targets. We refer to this phenomenon as \emph{error drift}, also referred to as \emph{confirmation bias}~\cite{confirmationbias}. It is highlighted on the left plot of Figure~\ref{fig:dilemma} where the ratio of correct and confident pseudo-label drops at some point when too many incorrect pseudo-labels were used in previous iterations.

\mypar{Signal scarcity.} Let $r_{\theta}(\tau)$ be the expected proportion of points that do not receive a pseudo-label when using Eq.~\ref{eq:distill}: 
\begin{equation} r_{\theta}(\tau) =\underset{\bx \sim P_x}{\mathbb{E}}  \llbracket \max \bm{\bar{p}_\theta}(\bx) < \tau \rrbracket .
\end{equation} For fixed model parameters $\theta$, $r_{\theta}(\tau)$ is monotonously increasing in $\tau$. With few ground-truth labels, most unlabelled images will be too dissimilar to any labeled one to obtain confident pseudo-labels early in training. Thus for high values of $\tau$, $r_\theta(\tau)$ will be close to $1$ and 
most data points masked by $\llbracket . \ge \tau \rrbracket$ in Equation~\ref{eq:distill}, thus
providing no gradient. The network receives scarce training signal; in the worst cases training will never start, or plateau early. We refer to this problem as \emph{signal scarcity}. This is illustrated in Figure~\ref{fig:dilemma} on the right plot where the ratio of images with confident pseudo-label remains low, meaning that many unlabeled images are actually not used during training.
    
\mypar{The distillation dilemma.}
	We now argue that the success of the FixMatch algorithm hinges on its ability to navigate the pitfalls of \emph{error drift} and \emph{signal scarcity}.
	Erroneous predictions, as measured by $P(E_{\theta}(\bx, \tau))$, are avoided by increasing the hyper-parameter $\tau$.
	Thus the set of values that avoid error drift can be assumed of the form $\nabla = [\tau_d, 1]$ for some $\tau_d \in [0, 1]$.
	Conversely avoiding signal scarcity, as measured by $r_{\theta}(\tau)$, requires reducing $\tau$, and the set of admissible values can be assumed of the form $\Delta = [0, \tau_s]$ for some $\tau_s \in [0,1]$.
	 Successful training with Equation~\ref{eq:distill} requires the existence of a suitable value of $\tau$, \ie, $\Delta \cap \nabla \neq \varnothing$, and that this $\tau$ can be found in practice. 
    On CIFAR-10 strikingly low amounts of labels are needed to achieve that \cite{fixmatch}. However we show that it is not the case on more challenging datasets such as STL-10, see Figure~\ref{fig:dilemma}.

\section{Proposed method}
\label{sec:method}

In this section, we introduce our method to overcome the distillation dilemma (Section~\ref{sub:method}) and introduce two improvements for selecting pseudo-labels 
(Section~\ref{sub:tricks}). 

\subsection{Alleviating the distillation dilemma}
\label{sub:method}
	
	In FixMatch, the absence of confident pseudo-labels leads to the absence of training signal, which is at odds with the purpose of consistency regularization - to allow training in the absence of supervision signal - and leads to the distillation dilemma. We propose instead to decouple self-training and consistency-regularization, by using self-supervision in case no confident pseudo-label has been assigned. While still relying on consistency regularization, the self-supervision does not depend at all on the labels or the classes, thus it is significantly different from previous works which use consistency regularization depending on the predicted class distribution of the weak augmentation to train the strong augmentation~\cite{remixmatch,uda}.

	\blue{When $\mathcal{L}_{\text{distill}}$ in Equation~\ref{eq:distill} does not provide training signal, we optimize consistency regularization (Eq.~\ref{eq:cr}) between strongly and weakly augmented images.
	Let $c_\theta(\bx_u) =	\llbracket \max \bm{p_\theta}(\bx_u) \ge \tau \rrbracket$ (equal to $1$ if the model makes a confident prediction, $0$ otherwise) we minimize:}
	\begin{equation}
	   	 \mathcal{L}_{\text{ours}}^\theta(\bx_u, \bx_s) = \Big[c_\theta(\bx_u) \cdot \mathcal{L}_{\text{distill}}^\theta + (1 - c_\theta(\bx_u)) \cdot
		\mathcal{L}_{\text{coreg}}^\theta\Big](\bx_u, \bx_s)\\
	    \label{eq:rich}
	\end{equation}

	By design, the gradients of this loss are never masked. Thus, in settings with hard data and scarce labels, it is possible to use a very high value for $\tau$, to avoid error-drift, without wasting most of the computations. In practice at each batch, images are sampled from $\mathcal{S}$ and $\mathcal{U}$, transformations from $\mathcal{A}_{\text{weak}}$, $\mathcal{A}_{\text{strong}}$ and we minimize:
	\begin{equation}
	    \hspace*{-0.1cm} \sum_{\bx_i \in \mathcal{S}} -\log p_{\theta}(y|w_i(\bx_i)) + \sum_{\bx_j \in \mathcal{U}} \mathcal{L}_{\text{ours}}^{\theta}(w_j(\bx_j),s_j(\bx_j)).
	\end{equation}

	 \blue{Consistency regularization is prone to collapse to trivial solutions. To avoid these solutions, we perform \textbf{online deep-clustering}, following~\cite{swav}. This solution is advantageous in terms of computational efficiency, as it does not require extremely large batch sizes~\cite{simclr}, storing a queue~\cite{moco}, or an exponential moving average model for training~\cite{byol}. 
	 Online deep-clustering works by projecting the images in a deep feature space and clustering them using the Sinkhorn-Knopp algorithm.
		Let $q_a$ a soft cluster assignment operator over $k$ classes, these are used as target for model predictions $q_{\theta}$ by predicting the assignment $q_a(\bx_u)$ of an augmentation $\bx_u$ from another augmentation $\bx_v$ and vice-versa, which yields the following consistency-regularization
		objective:}
		\begin{equation}
		\mathcal{L}_{\text{coreg}}^{\theta}(\bx_u, \bx_v) = \sum_{i=1}^k q_a^i(\bx_u)\log q_\theta^i(\bx_v) + q_a^i(\bx_v)\log q_\theta^i(\bx_u).
		\label{eq:clust}
		\end{equation}
		Because $q_a$ ensures that all clusters are well represented, the problem cannot be solved by trivial constant solutions. 
		Figure~\ref{fig:overview} gives an overview of our approach, \blue{where a pseudo-label is used on the strong augmentation if confident, and a feature cluster assignment is used otherwise}.
		
\mypar{Self-supervised pre-training.}
    An alternative 
    to leverage self-supervision is to use a \textit{self-then-semi} paradigm, \ie, to first pretrain the network using unlabeled consistency regularization, then continue training using FixMatch, as in~\cite{selfmatch}. 
    We hypothesize, and verify experimentally, that it is beneficial to optimize both simultaneously 
    rather than sequentially. Indeed, self-supervision yields representations that are not tuned to a specific task. Leveraging the information contained in ground-truth and pseudo-labels is expected to produce representations more aligned with the final task, which in turn can lead to better pseudo-labels.
    Empirically, we also find that self-supervised models transfer quickly but yield over-confident predictions after a few epochs, and thus suffer from strong error drift, see Section~\ref{sec:xp}.
\subsection{Improving pseudo-label quality}
Here we propose two methods to refine pseudo-labels  beyond thresholding $\text{softmax}$ outputs with a constant $\tau$.
\label{sub:tricks}

	\mypar{Method 1: Avoiding errors by estimating consistency.}
	\label{sec:consistency}
	As $\bm{p_\theta}(\bx)$ is used as confidence measure, the mass allocated
	to the class $c$ would ideally be equal to the probability of it being correct.
	Such a model is called \emph{calibrated}, formally defined as:
	\begin{equation} P(\argmax \bm{p_\theta(\bx)} = y) = p_\theta^y(\bx). \end{equation} 
	Unfortunately, deep models are notoriously hard to calibrate and strongly lean towards over-confidence 
	\cite{calibre_deep,measure_calibration,learnedsoft}, 
	which degrades pseudo-labels confidence estimates.
	At train time, augmentations come into play; Let $\mathcal{A}_{x, \theta}^c$ the set of transformations for which $\bx$ is classified as $c$:
	\begin{equation}
	    \mathcal{A}_{\bx, \theta}^c = \{u \in \mathcal{A} | \argmax \bm{p_{\theta}}(\bx_u) = c\}.
	\end{equation}
	The probability of $\bx$ being well classified by $\bm{p_\theta}$ is the measure: $\mu(\mathcal{A}_{\bx, \theta}^y)$ with $y$ the true label. 
	For unlabeled images, this cannot be estimated empirically as $y$ is unknown.
	Instead we use prediction consistency as proxy: assume the most predicted class $\hat{y}$ is correct\footnote{\tiny This proxy can also lead to error drift, but the confidence test is designed to be very stringent.
} and estimate $\mu(\mathcal{A}_{\bx, \theta}^{\hat{y}}).$
	Empirically, we are interested in testing the hypothesis:
	$$h: `(\mu(\mathcal{A}_{\bx, \theta}^{\hat{y}}) \ge \lambda)' \text{ with confidence threshold }\alpha.$$ 
	Note that for any class $c$,  $(\mu(\mathcal{A}_{\bx, \theta}^{c}) \ge 0.5)$  implies $\hat{y} = c$. Hypothesis $h$ can be tested with a Bernoulli parametric test: let $\hat{\mu}_{\bx, \theta}^c$ be the empirical  estimate of $\mu(\mathcal{A}_{\bx, \theta}^{c})$; We are interested in $\hat{\mu}_{\bx, \theta}^c$ close to $1$, so assuming $N \ge 30$, $[\hat{\mu}_{\bx, \theta}^c - 3/N; 1]$ is approximately a $95\%$-confidence interval \cite{ruleof3}.	In practice, we amortize the cost of the test by accumulating a history of predictions 
	for $\bx$, of length $N$, at different iterations; there is a trade-off between how stale the predictions are and the number of trials. At the end of each epoch, data points that pass our approximate test for $h$ are added to the labeled set, for the next epoch. 

\mypar{Method 2: Class-aware confidence threshold.}
    \label{sec:aware}
    The optimal value for the confidence threshold $\tau$ in Equation~\ref{eq:rich} depends on the model prediction accuracy. 
    In particular, different values for $\tau$ can be optimal for different classes and at different times. Classes that rarely receive pseudo-labels may benefit from more `curiosity' with a lower $\tau$, while classes receiving a lot of high quality labels may benefit from being conservative, with a higher $\tau$. To go beyond a constant value of $\tau$ shared across classes we assume that an estimate $r_c$ of the proportion of images in class $c$, is available\footnote{\tiny This is a very mild assumption: it is sufficient to assume that labels are sampled in an i.i.d. manner, in which case empirical ratios are unbiased estimates of $r_c$ -- though possibly high variance when labels are scarce. In any case on CIFAR-10/100 and STL-10, the standard protocol \cite{mixmatch,fixmatch}, which we follow, makes a much stronger assumption: images are sampled uniformly across classes, which ensures that $r_c$ is known exactly for all $c$.} and estimate $p_c$ the proportion of images confidently labeled into class $c$.
    At each iteration we perform the following updates: 
    \begin{eqnarray}
        p_c^{t+1} &=& \alpha p_c^{t} + (1 - \alpha) p_c^{\text{batch}} \\
    \label{eq:tauupdate}
        \tau_c^{t+1} &=& \tau_c^{t} + \epsilon \cdot \text{sign}(p_c - r_c)
    \label{eq:pcupdate}
    \end{eqnarray}
    Equation~\ref{eq:tauupdate} decreases $\tau_c$ for classes that receive less labels than expected, to explore uncertain classes more. Conversely, the model can focus on the most certain images for well represented classes. This procedure introduces two hyper-parameters ($\alpha$ and $\epsilon$), but these only impact how fast $\tau$ and $p_c$ are updated. In practice we did not need to tune them and used default values of $\alpha=0.9$ and $\epsilon=0.001$.
\section{Experimental results}
\label{sec:xp}

\begin{figure*}
 \centering
 \includegraphics[width=0.8\linewidth]{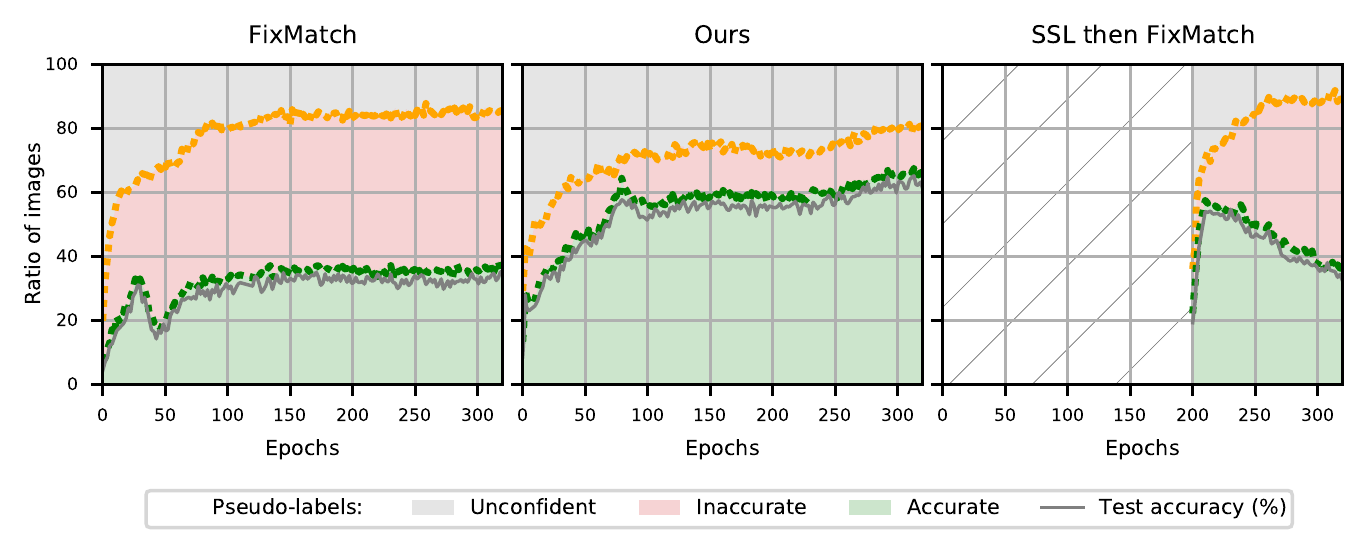} \\
  \mvspace{-0.1cm}
  \vspace*{-0.2cm}
 \caption{Evolution of the pseudo-labels during training. For each method (FixMatch, our composite loss, and SSL-then-FixMatch from left to right), we show the ratio of images with correct and confident pseudo-labels (bottom green area), incorrect but confident pseudo-labels (red area) and unconfident pseudo-labels (grey area). We also plot the test accuracy with a black line. For SSL-then-FixMatch, early training corresponds to self-supervised learning, thus these information are not available.}
 \label{fig:mainanalysis}
 \mvspace{-0.4cm}
\end{figure*}

We present the datasets and experimental setup in Section~\ref{sub:setup} and validate our general idea on STL-10 in Section~\ref{sub:validate}. We then ablate the improvements for the pseudo-label accuracy in Section~\ref{sub:xppseudolabel}, add evaluations on CIFAR in Section~\ref{sub:cifar} and compare to the state of the art in Section~\ref{sub:sota}.

\subsection{Experimental setup}
\label{sub:setup}
\begin{figure*}
\hspace*{-1cm}

\centering
	\includegraphics[width=0.33\linewidth]{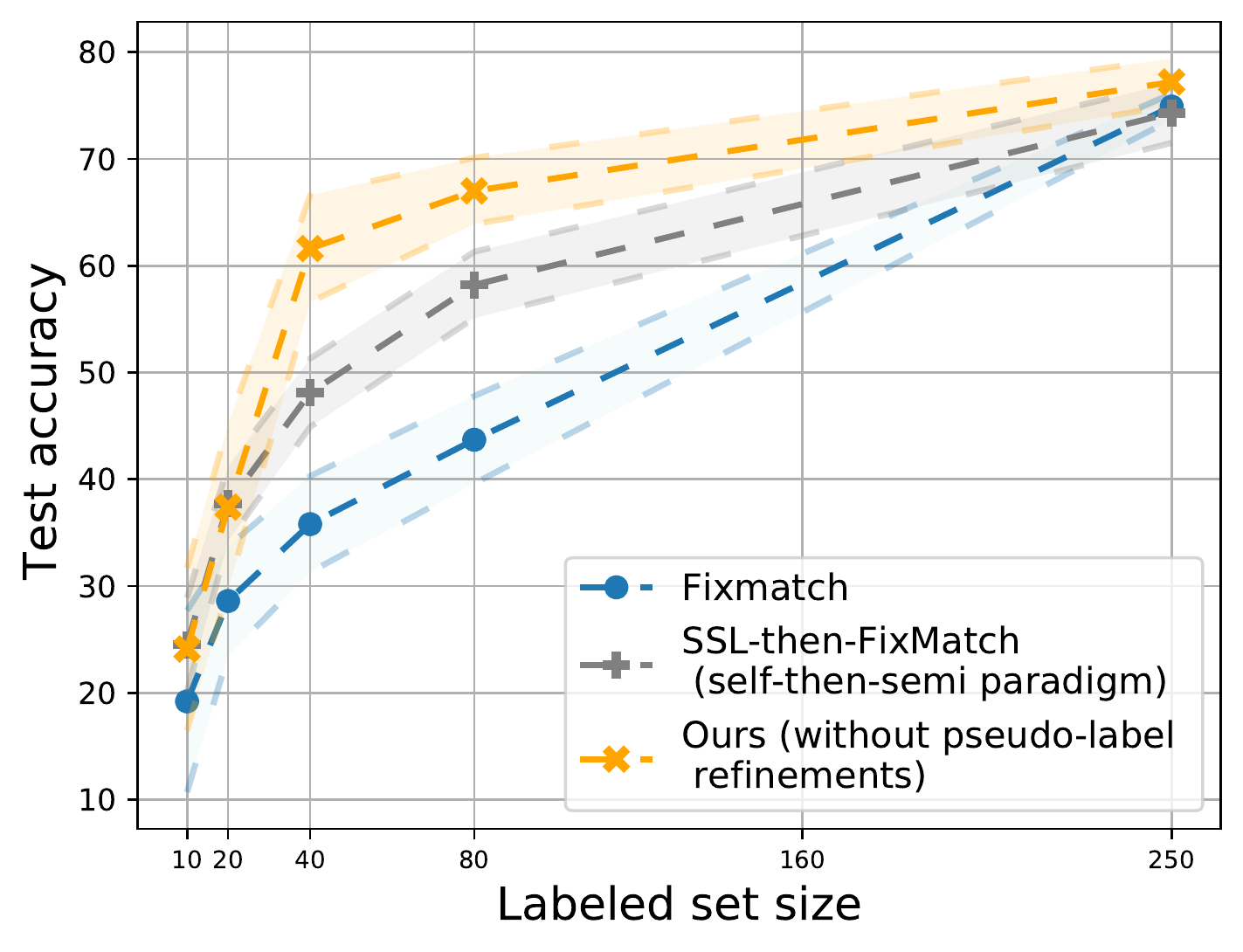}%
	\includegraphics[width=0.32\linewidth]{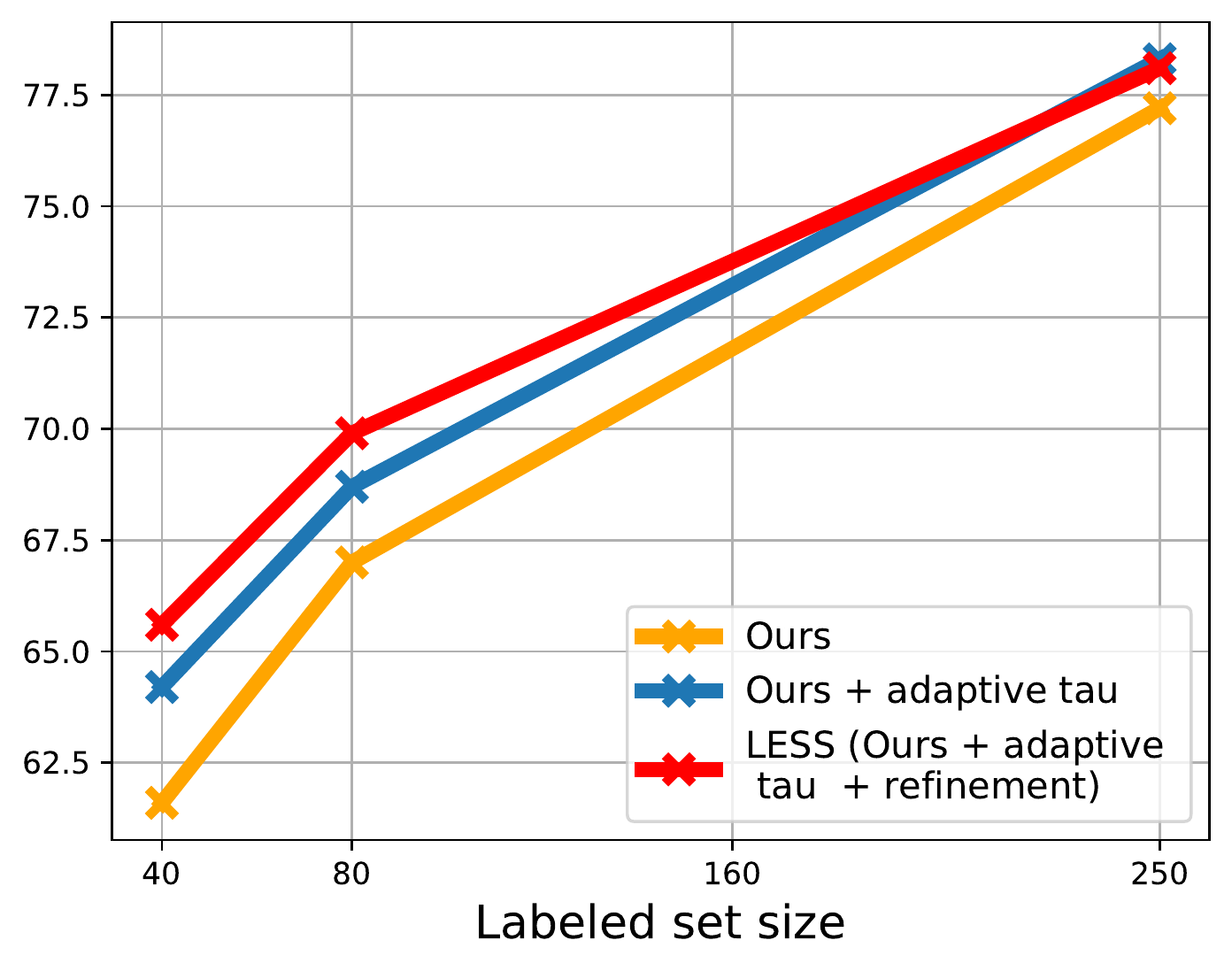}%
	\begin{minipage}{0.33\linewidth}
	\vspace{-4.5cm}
\resizebox{\linewidth}{!}{
\vspace*{-2cm}
\renewcommand{\arraystretch}{1.7}
\setlength{\tabcolsep}{3pt}
\begin{tabular}{cccc}
\toprule
 &  \multicolumn{3}{c}{STL-10}  \\
&$40$ labels & $80$ labels & $250$ labels\\
\cmidrule(lr){1-1} \cmidrule(lr){2-4}
Fixmatch &  $35.8$ \scriptsize{}{}{$\pm 4.5$} & $43.7$ \scriptsize{}{}{$\pm 4.1$}  & $74.9$ \scriptsize{}{}{1.2}  \\
SSL-then-FixMatch &  $48.1$ \scriptsize{}{}{$\pm 3.2$} & $58.2$ \scriptsize{}{}{$\pm 2.1$}  & $74.3$ \scriptsize{}{}{$\pm 2.8$}  \\
 {\renewcommand{\arraystretch}{0.6}
\begin{tabular}{@{}c@{}} Ours \footnotesize{(no adaptive} \\ \footnotesize{$\tau$ \& no refinement)}\end{tabular}} &  $61.6$ \scriptsize{}{}{$\pm 5.0$} & $67.0$ \scriptsize{}{}{$\pm 3.1$}  & $77.2$ \scriptsize{}{}{$\pm 2.2$}  \\
{\renewcommand{\arraystretch}{0.6}
\begin{tabular}{@{}c@{}}Ours  \\ \footnotesize{(no refinement)}\end{tabular}} &  $64.2$ \scriptsize{}{}{$\pm 5.1$} & $68.7$ \scriptsize{}{}{$\pm 3.0$}  & $\bf{78.3}$ \scriptsize{}{}{$\pm 2.3$}  \\
{\renewcommand{\arraystretch}{0.6}
\begin{tabular}{@{}c@{}}\tabours \: \footnotesize{(Ours} \\ \footnotesize{ + adaptive $\tau$ + refinement)} \end{tabular}}&  $\bf{65.6}$ \scriptsize{}{}{$\pm 5.0$} & $\bf{69.9}$ \scriptsize{}{}{$\pm 3.1$}  & $78.1$ \scriptsize{}{}{$\pm 2.2$}  \\
\bottomrule
\end{tabular}\vspace{5cm}
}
\end{minipage}\\
\begin{flushleft}
{\small \vspace*{-0.15cm} \hspace{3cm}  (a) \hspace{5cm} (b) \hspace{5.5cm} (c)}
\end{flushleft}


%



	\caption{(a) Classification accuracy on STL-10 for various sizes of labeled sets; standard deviations represented with light colors. Substantial gains are observed compared to FixMatch and the self-then-semi paradigm.
	(b) Results on STL-10 for various sizes of labeled set, with our two improvements to refine pseudo-labels.
	(c) summary of our main results on STL-10.
	}

	\label{fig:main}
	

\end{figure*}

We perform most ablations on STL-10 and also compare approaches on CIFAR-10 and CIFAR-100. 
We rely on a wide-ResNet WR-28-2~\cite{wideresnet} for CIFAR-10, WR-28-8 for CIFAR-100 and WR-37-2 for STL-10 and follow FixMatch for data augmentations.

\mypar{The STL-10 dataset} consists of 
$5$k labeled images 
of resolution $96\times96$
split into $10$ classes, and 
$100$k unlabeled images. It contains images with significantly more variety and detail than images in the CIFAR datasets; it is extracted from ImageNet and unlabeled images can be very different from those in the labeled set. It remains manageable in terms of size, with twice as many images as in CIFAR-10, offering an interesting trade-off between challenge and computational resources required.
We use various amounts of labeled data: $10$ ($1$ image per class), $20$, $40$, $80$, $250$, $1000$. 



\mypar{Metric.} We report top-1 accuracy for all datasets. In barely-supervised learning, the choice of the few labeled images can have a large impact on the final accuracy, so we report means and standard deviations over multiple runs. Standard deviations increase as the number of labels decreases, so we average across $4$ random seeds for $4$ images per class or less, $3$ otherwise, and across the last $10$ checkpoints of all runs.
\mvspace{-0.2cm}

\subsection{\hspace*{-0.15cm}Validating our composite approach on STL-10}
\label{sub:validate}

To validate our approach, we train the baselines and our models with progressively smaller sets of labeled images; the goal is to reach a performance that degrades gracefully when progressively going to the barely-supervised regime.

To demonstrate the benefit of our composite loss from Equation~\ref{eq:rich} (without the proposed pseudo-label quality improvements of Section~\ref{sub:tricks}), we first compare this to the original FixMatch loss in Figure~\ref{fig:main} (c) on the STL-10 dataset when training with different sizes of labeled sets, namely $\{10, 20, 40, 80, 250\}$ labeled images. We use $\tau=0.95$ for FixMatch and $\tau=0.98$ for our model, see Section~\ref{sub:xppseudolabel} for discussions about setting $\tau$.
Our approach (yellow curve) significantly outperforms FixMatch (blue curve), especially in the regime with $40$ or $80$ labeled images where the test accuracy improves by more than $20\%$. When more labeled images are considered (\eg  $250$), the gain is smaller. When only $1$ image per class is labeled, the difference is also small but our approach remains the most label efficient.

We also compare to a method using a \textit{self-then-semi} paradigm, where online deep-clustering is first used alone before FixMatch is run on top of this pretrained model (grey curve labeled Self-then-FixMatch). While it performs better than FixMatch applied from scratch, we also outperform this approach, in particular in barely-supervised scenario, \ie, with less than 10 images per class.

To better analyze these results, we show in Figure~\ref{fig:mainanalysis} the evolution of pseudo-label quality for our approach, FixMatch and SSL-then-FixMatch. 
Our method has less examples with confident pseudo-labels in the early training; we can set a higher value of $\tau$, as we do not suffer from signal scarcity in case of unconfident pseudo-labels.
In contrast, FixMatch assigns more confident pseudo-labels in early training, at the expense of a higher number of erroneous pseudo-labels, leading eventually to more errors due to error drift, also named confirmation bias.
Note that the test accuracy is highly correlated to the ratio of training images with correct pseudo-labels, and thus error drift harms final performance.
When comparing SSL-then-FixMatch to FixMatch, we observe that the network is quickly able to learn confident predictions, with a lesser ratio of incorrect pseudo-labels. However this ratio is still higher than with our approach. When evaluating pre-trained models, we use model checkpoints obtained between $10$ and $20$ epochs, because more training harms the performance due to confirmation bias. This was cross-validated on a single run using $80$ labeled images, and used for all other seeds and labeled sets.

\subsection{Improving pseudo-label accuracy}
\label{sub:xppseudolabel}

We now evaluate the modifications proposed in Section~\ref{sub:tricks}, as well as the impact of $\tau$ to further improve performance. 

\noindent \textbf{Impact of the confidence threshold.} The simplest way to trade-off quality and the amount of pseudo-labels, both for FixMatch and our method, is to change the confidence threshold. In Table~\ref{tab:tau}, we train both methods for values of $\tau$ in $\{0.95, 0.98, 0.995\}$, with labeled-split sizes in $\{40, 80\}$. The first finding is that the average performance of FixMatch degrades when increasing $\tau$; in particular, with $40$ labeled images, it drops by $1.9\%$ (resp. $7.4\%$) when increasing $\tau$ from $0.95$ to $0.98$ (resp. $0.995.$)
Thus the default value of $\tau=0.95$ used in~\cite{fixmatch}, is the best choice; the improved pseudo-label quality obtained from increasing $\tau$ is counterbalanced by signal scarcity, see Section~\ref{sub:dillema}. 

On the other hand, the performance of our method improves when increasing $\tau$; in particular, with $40$ labeled images, it increases by $2.4\%$. 
As expected, our method benefits from using self-supervised training signal in the absence of confident pseudo-labels, which allows us to raise $\tau$
without signal scarcity, and  without degrading the final accuracy. The performance of our method remains stable when raising $\tau$ to $0.995$; this demonstrates that it is robust to high threshold values, even though this does not bring further accuracy improvements. For the rest of the experiments we keep $\tau=0.98$ for our method and $\tau=0.95$ for FixMatch.

\begin{table}
 \centering
 \resizebox{\linewidth}{!}{
\begin{tabular}{ccccc}
\toprule
 \multirow{2}{*}{$\tau$}      & \multicolumn{2}{c}{FixMatch} & \multicolumn{2}{c}{Ours \footnotesize{}{}{(w/o refinements)}} \\
 & 40 labels & 80 labels & 40 labels & 80 labels \\
\cmidrule(lr){1-1} \cmidrule(lr){2-3} \cmidrule(lr){4-5}
0.95    & $\bf{35.8}$ \footnotesize{}{}{$\pm 4.5$} &  $\bf{48.1}$  \footnotesize{}{}{$\pm 3.2$} & $61.8$ \footnotesize{}{}{$\pm 4.9$} & $67.2$ \footnotesize{}{}{$\pm 2.9$} \\
0.98    & $33.9$ \footnotesize{}{}{$\pm 4.7$} & $47.4$ \footnotesize{}{}{$\pm 3.5$} & $\bf{64.2}$ \footnotesize{}{}{$\pm 5.1$} & $\bf{68.7}$ \footnotesize{}{}{$\pm 3.0$} \\
0.995   & $28.4$ \footnotesize{}{}{$\pm 6.2$} & $46.3$ \footnotesize{}{}{$\pm 3.7$} & $64.1$ \footnotesize{}{}{$\pm 4.3$} & $68.6$ \footnotesize{}{}{$\pm 2.8$} \\
\bottomrule
\end{tabular}
 }
 
 	\mvspace{-0.3cm}

	\caption{\small Ablation on the threshold parameter $\tau$ on the STL-10 dataset for $40$ and $80$ labeled images. The default value of $0.95$ works best for FixMatch. Our method benefits from increasing $\tau$ to $0.98$, and is more robust to a higher threshold value of $0.995$.}
	\label{tab:tau}
	\mvspace{-0.4cm}
\end{table}

\mypar{Adaptive threshold and confidence refinement.} 
We now validate the usefulness of the class-aware confident threshold presented in Section~\ref{sec:aware}. We plot in Figure~\ref{fig:main} (b) the performance of our model, with (blue line) and without it (yellow line). Adaptive thresholds demonstrate consistent gains across labeled-set sizes, \eg with an average gain of $2.6\%$ when using $40$ labels. This validates the approach of bolstering the exploration of classes that are under represented in the model predictions, while focusing on the most confident labels for classes that are well represented. The gains observed are more substantial for low numbers of labeled images, like $40$ compared to $250$, which suggests that when using a fixed threshold, exploration may naturally be more balanced with more labeled images.


\mypar{Impact of using pseudo-label refinement on our method.}  
We now evaluate the refinement of pseudo-labels using a set of predictions for different augmentations $u \in \mathcal{A}$, see Section~\ref{sec:aware}. Figure~\ref{fig:main}(b) reports performance for models trained with Equation~\ref{eq:rich}, with (red curve) and without refined labels (blue curve). Using the refined labels offers a $1.4\%$ (resp. $1.1\%$) accuracy improvement on average when using $40$ (resp. $80$) labels, on top of the gains already obtained from using our composite loss, and adaptive thresholds. No improvement is observed, however, with 250 labels.



\subsection{Comparison on CIFAR}
\label{sub:cifar}

\begin{table}
\centering
\setlength{\tabcolsep}{3pt}
\resizebox{1.\linewidth}{!}{
\begin{tabular}{cccc|ccc}
\toprule
 &  \multicolumn{3}{c|}{CIFAR-10} & \multicolumn{3}{c}{CIFAR-100} \\
&$10$ labels & $40$ labels & $250$ labels & $100$ labels& $200$ labels & $400$ labels \\
\cmidrule(lr){1-1} \cmidrule(lr){2-4} \cmidrule(lr){5-7}
FixMatch & $56.1$ \scriptsize{}{}{$\pm 11.3$} & $92.1$ \scriptsize{}{}{$\pm 3.4$}  & $94.0$ \scriptsize{}{}{$\pm 0.9$} & $23.1$ \scriptsize{}{}{$\pm 4.7$}  & $38.6$ \scriptsize{}{}{$\pm 3.5$} & $50.2$ \scriptsize{}{}{$\pm 2.1$}  \\
\tabours &  $\bf{64.4}$ \scriptsize{}{}{$\pm 10.9$} & $\bf{93.2}$ \scriptsize{}{}{$\pm 2.1$}  & $\bf{95.0}$ \scriptsize{}{}{$\pm 0.8$} & $\bf{28.2}$ \scriptsize{}{}{$\pm 3.0$} & $\bf{42.5}$ \scriptsize{}{}{$\pm 3.2$} & $\bf{51.3}$ \scriptsize{}{}{$\pm 2.4$}  \\
\bottomrule
\end{tabular}
}
\mvspace{-0.3cm}
\caption{Comparison of our approach and FixMatch for barely-supervised learning on CIFAR-100 and CIFAR-10.
All results obtained using the same code base. As labels becomes more scarce, greater performance gains are observed with our method. }
\label{tab:cifar}
\mvspace{-0.1cm}
\end{table}

So far, we drew the comparison on STL-10 dataset only. We now compare our method with pseudo-labels quality improvements, denoted as \textbf{LESS} for Label-Efficient Semi-Supervised learning, to FixMatch on CIFAR-10 and CIFAR-100 with labeled set sizes of $1$, $2$ or $4$ samples per class in Table~\ref{tab:cifar}. For CIFAR-10 we report results for our model with $\tau=0.995$, as we found it to be the best among $\{0.95, 0.98, 0.995\}$. We observe that our approach outperforms FixMatch for all cases, with a gain ranging from $5\%$ with 1 label per class to $1\%$ with 4 labels per class on CIFAR-100, and from $8\%$ with 1 label per class to $1\%$ with 25 labels per class on CIFAR-10. The gain here is smaller than the one reported on STL-10. We hypothesize that the very low resolution ($32 \times 32$) of CIFAR images leads to less powerful self-supervised training signals.


\subsection{Comparison to the state of the art}
\label{sub:sota}


We finally compare our approach to numbers reported in others papers in Table~\ref{tab:sota}. Note that all previous papers reported numbers on STL-10 are with 1000 labels (\ie, 100 labels per class), which cannot be considered as barely-supervised.
We thus only provide a comparison on the CIFAR-10 dataset where other methods reported results for smaller numbers of images per class. 
We observe that our approach performs the best on the CIFAR-10 dataset, in particular with 4 labels per class. 
Tthe gap with the self-then-semi paradigm is lower on this dataset as it is significantly less challenging than STL-10.
Note that the previous results of FixMatch in Table~\ref{tab:cifar} were obtained with our code-base, which explains the slightly different performance compared to the numbers in Table~\ref{tab:sota}.

	\begin{table}
	    \setlength{\tabcolsep}{7pt}
	  \centering
	  \resizebox{\linewidth}{!}{
	  \begin{tabular}{lll}
	    \toprule
	     & \multicolumn{2}{c}{CIFAR-10} \\
	    
	    
	                 & 40 labels & 250 labels \\

	    \cmidrule(lr){1-1} \cmidrule(lr){2-3} 
	    
	    \multicolumn{3}{l}{\textit{Semi-supervised from scratch}} \\
	    
	    ~~~~Pseudo-Label~\cite{pseudolabel}   & - & $50.2$ \footnotesize{}{}{$\pm0.4$} \\
	    
	    ~~~~$\Pi$-Model~\cite{rasmus2015semi}                & - & $45.7$\footnotesize{}{}{$\pm 3.9$} \\ 
	    ~~~~Mean Teacher~\cite{meanteacher}                & - & $67.7$\footnotesize{}{}{$\pm 2.3$} \\ 
	    

	    ~~~~MixMatch~\cite{mixmatch}		            & $52.5$ \footnotesize{}{}{$\pm 11.5$} & $89.0$ \footnotesize{}{}{$\pm 0.9$} \\ 

	    ~~~~UDA~\cite{uda}                         & $71.0$ \footnotesize{}{}{$\pm 5.9$} & $91.2$ \footnotesize{}{}{$\pm 1.1$} \\
	    ~~~~ReMixMatch~\cite{remixmatch}               & $80.9$ \footnotesize{}{}{$\pm 9.6$} & $94.6$ \footnotesize{}{}{$\pm 0.1$} \\
	    ~~~~FixMatch~\cite{fixmatch} (with RA~\cite{randaugment})     	        & $86.1$ \footnotesize{}{}{$\pm 3.4$} & $94.9$ \footnotesize{}{}{$\pm 0.7$} \\

	    
	    \noalign{\smallskip}
	    \multicolumn{3}{l}{\textit{Self-then-semi paradigm}} \\
	    
	    ~~~~SelfMatch~\cite{selfmatch} & $\bf{93.2}$ \footnotesize{}{}{$\pm 1.1$} & $\bf{95.1}$ \footnotesize{}{}{$\pm 0.3$} \\
	    ~~~~CoMatch~\cite{comatch} & $93.1$\footnotesize{}{}{$\pm 1.4$} & $\bf{95.1}$ \footnotesize{}{}{$\pm 0.3$} \\
	    
	    
	    \noalign{\smallskip}
	    \multicolumn{3}{l}{\textit{Composite self- and semi-supervised}}\\
	    
	    ~~~~\textbf{\tabours} & $\bf{93.2}$ \footnotesize{}{}{$\pm 2.1$} & $\bf{95.1}$ \footnotesize{}{}{$\pm 0.8$} \\
	    
	    \cmidrule(lr){1-1} \cmidrule(lr){2-3} 
	    	    \textit{Fully-Supervised}                  & \multicolumn{2}{c}{95.9} \\
	    	    \bottomrule
	    
	  \end{tabular}
	  }
     \mvspace{-0.3cm}
	  \caption{Comparison to state-of-the-art methods on CIFAR.}
      \label{tab:sota}
      \mvspace{-0.5cm}
  	\end{table}

\section{Conclusion}

After analyzing the behavior of FixMatch in the barely-supervised learning scenario, we found that one critical limitation was due to the distillation dilemma. We proposed to leverage self-supervised training signals when no confident pseudo-label were predicted and showed that this composite approach
allows to significantly increase performance. We additionally proposed two refinement strategies to improve pseudo-label quality during training and further increase test accuracy.
Further research directions include extension to datasets with more classes such as ImageNet. 
Other related topics such as model calibration and learning with noisy labels are also directions that we expect to be critical to progress in barely-supervised learning.
\newpage

\bibliography{ref}

\end{document}